# Legal Decision-making for Highway Automated Driving

Xiaohan Ma*, Wenhao Yu*. Chengxiang Zhao, Changjun Wang ⊠, Wenhui Zhou, Guangming Zhao, Mingyue Ma, Weida Wang, Lin Yang, Rui Mu, Hong Wang ⊠, IEEE senior member, Jun Li

*Abstract*— **Compliance with traffic laws is a fundamental requirement for human drivers on the road, and autonomous vehicles must adhere to traffic laws as well. However, current autonomous vehicles prioritize safety and collision avoidance primarily in their decision-making and planning, which will lead to misunderstandings and distrust from human drivers and may even result in accidents in mixed traffic flow. Therefore, ensuring the compliance of the autonomous driving decision-making system is essential for ensuring the safety of autonomous driving and promoting the widespread adoption of autonomous driving technology. To this end, the paper proposes a trigger-based layered compliance decision-making framework. This framework utilizes the decision intent at the highest level as a signal to activate an online violation monitor that identifies the type of violation committed by the vehicle. Then, a four-layer architecture for compliance decision-making is employed to generate compliant trajectories. Using this system, autonomous vehicles can detect and correct potential violations in real-time, thereby enhancing safety and building public confidence in autonomous driving technology. Finally, the proposed method is evaluated on the DJI AD4CHE highway dataset under four typical highway scenarios: speed limit, following distance, overtaking, and lane-changing. The results indicate that the proposed method increases the vehicle's overall compliance rate from 13.85% to 84.46%, while reducing the proportion of active violations to 0%, demonstrating its effectiveness.**

## I. INTRODUCTION

WITH the advancement of the automotive industry, road traffic safety issues are receiving a growing amount of attention. *The Statistical Annual Report of Road Traffic Accidents of the People's Republic of China* reveals that between 2017 and 2020, China had approximately 235,100 road accidents and 62,900 deaths per year, with approximately 86.6% of these accidents being the result of drivers violating traffic laws. According to the University of Michigan Traffic Research Institute in the United States, automated driving technology has the potential to prevent 85% of traffic accidents. At the same time, the European Union has presented a vision for sustainable transportation that aims for zero fatalities by 2050[1]. It is reasonable to assume that the development of autonomous vehicles (AVs) will substantially reduce accident rates and enhance overall traffic safety[2]. However, with the development of AVs, AVs and human-driven vehicles will inevitably share the road in the foreseeable future, which poses a greater challenge to the safety of AV decision-making systems[3]. Unreasonable decision-making can cause traffic congestion and even result in fatal accidents.

To ensure the safety of AVs in traffic with human-driven vehicles, it is essential that AVs adhere to the same traffic laws as human drivers. Traffic laws are not only an important guarantee of safety and the cornerstone of unmanned landing but also a critical foundation for subsequent accident liability division[4]. On the other hand, given the relatively simple driving conditions on highways, it is expected that they will be one of the first application scenarios of autonomous driving technology. However, reports of accidents involving AVs on highways have increased significantly in recent years. Due to the typically high speeds on highways, accidents pose a direct risk to human life and property. Therefore, achieving compliance judgment and decision-making for AVs in highway scenarios is a key problem to be solved in the field of automated driving. However, the fuzzy nature of traffic law definitions and the slow development of decision-making system solutions have made it difficult to ensure that AVs comply with traffic laws for human drivers. Therefore, this paper focuses on legal decision-making for autonomous highway driving.

### A. Related research

Most of the existing decision-making systems for AVs only consider collision-free driving as a safety condition or simply add some parameter settings that take into account traffic laws. Few studies focus specifically on compliance with traffic laws in the decision-making process. Currently, there are two main schemes for decision-making considering traffic laws, that is, the rule-based method and the learning-based method[5]. There

The research presented in this paper is financially supported by the National Natural Science Foundation of China Project under Grant No. 52072215 and U1964203 and the National key R&D Program of China: 2022YFB2503003. (Corresponding author: Changjun Wang and Hong Wang)

Xiaohan Ma, Chengxiang Zhao, Weida Wang, and Lin Yang are with the school of mechanical engineering, Beijing Institute of Technology, Beijing 100081 China (e-mail: 3120210314@bit.edu.cn; 3220210327@bit.edu.cn;wangwd0430@163.com; yanglin@bit.edu.cn).

Wenhao Yu, Hong Wang, and Jun Li are with the school of Vehicle and Mobility, Tsinghua University, Beijing, 100084, China (e-mail: wenhaoyu@mails.tsinghua.edu.cn; hong_wang@mail.tsinghua.edu.cn; lj19580324@126.com).

Changjun Wang, Wenhui Zhou, Guangming Zhao and Mingyue Ma are with the Traffic Management Research Institute of the Ministry of Public Security, Wuxi , 214151 China (e-mail: wcj121@sina.com; 1983zhouwenhui, zgmbright@163.com; mmy1986@foxmail.com).

Rui Mu is with the school of automotive and traffic engineering, Jiangsu University, Suzhou, 212013, China (e-mail: 2212104062@stmail.ujs.edu.cn)

are primarily three methods for considering traffic laws in rule-based decision-making methods: constraint-based methods, finite state machine-based methods, and optimization-based methods. Constraint-based methods avoid violating laws to some extent by defining law-related parameters as state constraints or planning space constraints. For instance, Christian et al. [6]–[8] defined the concept of legal safety by combining traffic laws and safety, utilized reachable sets for prediction [9], [10], and generated feasible legal safety zones for collision-free driving on complex roads. However, this method combines the concept of traffic laws with safety and cannot guarantee completely compliant driving. In contrast, methods based on finite-state machines can predefine a set of compliant actions for corresponding scenarios, thereby achieving full compliance in particular scenarios. For instance, Lu et al. [11] proposed a method for generating candidate driving behavior sets based on traffic laws; however, this method is limited to a few specific scenarios and focuses primarily on static traffic laws without interaction. Optimization-based methods consider laws in decision-making planning by adding relevant traffic laws parameters to the planning cost function. For instance, Klemens et al. [12] proposed a game-based planning method that uses violations and collisions as reward functions for multi-objective interactive behavior planning. In addition, some scholars have considered vehicle-road coordination solutions, such as the use of a policy-based framework to define and allocate temporary traffic laws through roadside modules [13]. However, all these methods indirectly make the decision-making system consider traffic laws. To make the consideration of traffic laws more intuitive for AVs during driving, some scholars have started to consider digitizing traffic laws into logical languages. For instance, Cristian et al. [14] quantified passenger demand, safety preferences, and traffic laws using linear temporal logic (LTL) language and proposed a function to measure violations, unsafety, traffic safety, and traffic safety in order to resolve conflicts between road rules and passenger demand and safety preferences. Jana et al. [15] transformed traffic laws into Finite linear temporal logic (FLTL), defined the unsafety of vehicle trajectories based on the number of steps and priority weights violated in the violation of traffic laws, and then minimized the unsafety of the trajectory. However, this method is limited to static, non-interactive traffic laws, such as obstacle avoidance and driving in a particular direction, and compliance is sometimes sacrificed in order to increase traffic efficiency. Therefore, the current rule-based compliance decision-making methods primarily consider non-interactive or simple laws, such as speed limits and safe distances for lane changes, and have not yet developed a systematic compliance decision-making method.

With the rise of deep learning, some scholars have begun to consider using learning-based methods for decision-making and planning, with deep reinforcement learning being the most popular technique. A common way to consider traffic laws in deep reinforcement learning approaches is to design traffic laws as a constraint or reward functions and use them to train policies. By incorporating traffic laws as constraints, a policy that can take into account simple traffic laws can be trained by solving constrained Markov decision problems, such as methods based on Lagrange multipliers [16],[17], constrained strategy optimization methods [18], [19], etc. Besides, it is also a common method to consider the punishment for violation of traffic laws in the reward function [20], [21]. In addition to the above two methods, some scholars consider combining reinforcement learning with rule-based methods to make compliance decisions [22], such as Liu et al. [23], who proposed a reinforcement learning-based method with a rule-based backup policy to address the law adaptive decision-making problem. However, the learning-based decision-making methods are mostly end-to-end [24]. Due to the black-box nature of deep reinforcement learning, it remains to be determined if it can meet the high-reliability requirements of the auto-drive system.

### B. Contributions

Currently, most compliance decision-making methods only consider non-interactive traffic laws, such as road markings and traffic lights, which do not involve interactions. There is limited research on compliance decision-making methods for dynamic traffic laws involving interactions between vehicles and their surroundings. This paper proposes a trigger-based layered compliance decision-making method for dynamic traffic laws on highways with mixed traffic flow. This method can generate compliant trajectories by monitoring violations in real-time, thereby preventing vehicles from violating traffic laws. The main contributions of this paper are as follows:

- A trigger-based, layered compliance decision-making framework is proposed, based on the digitization of traffic laws, that can generate compliant trajectories tailored to the type of violation committed by the vehicle, thereby effectively preventing violations during vehicle operation.
- A method is proposed based on the classification of traffic laws to decompose traffic laws into basic violation types. Furthermore, a method for defining the priority of traffic laws in the event of conflicts is introduced, allowing for expedited compliance when multiple violations occur simultaneously.
- A compliance state transition strategy is proposed for the basic violation types of traffic laws, which can generate compliant references (solid red line, Fig.1) and compliant constraints (yellow dotted line, Fig.1) when the vehicle is in or about to be in a violation state, allowing the vehicle to return to compliant state quickly.

### C. Paper Organization

This paper is organized as follows. Section II introduces the trigger-based, layered decision-making framework for compliance. Section III introduces the method of decomposing traffic laws into basic violation types, as well as the method of generating compliance constraints and references, the setting of priority, and the compliance trajectory optimization method. Section IV introduces the simulation verification performed on the DJI AD4CHE dataset [25] in China to demonstrate the effectiveness of the proposed compliance decision-making method. In Section V, the conclusion and future work prospects are presented.

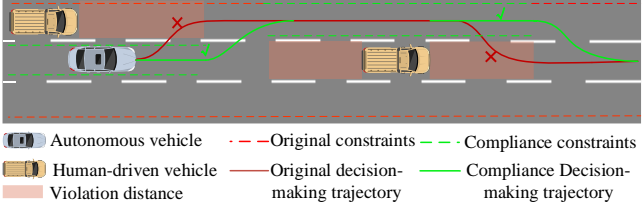

Fig. 1. Legal compliance decision-making process.
The original decision-making trajectory (solid red line) is planned by a motion planning planner that only considers collision avoidance while providing the initial constraints (red dotted line). When the state of the vehicle is predicted to violate traffic laws, our compliance decision-making method generates a new trajectory (solid green line) that causes the vehicle to abandon the original violation while generating compliance constraints (green dotted line) that prevent the vehicle from violating.

## II. TRIGGER-BASED LAYERED COMPLIANCE DECISION-MAKING FRAMEWORK

In this paper, a trigger-based, layered decision-making framework for autonomous vehicle compliance with highway traffic laws is proposed. To ensure the generality and scalability of the compliance decision-making method, the proposed approach is not integrated into the decision-making algorithm itself but constitutes a separate module after the motion planning layer. The proposed method achieves non-intrusive compliance decision-making using Model Predictive Control (MPC) decision-making algorithms by adding compliance reference and state constraints.

As shown in Fig.2, this paper employs the behavioral layer's decision intents as trigger signals for our framework. When a corresponding intent is generated, the compliance of the vehicle's state is monitored. This paper monitors the predicted state of the vehicle, which enables the generation of violation information when the vehicle is about to engage in noncompliant behavior. This information regarding violations is then transmitted to the compliance decision-making module. The motion planning layer generates a detailed reference trajectory, referred to in this paper as the initial reference trajectory.

The decision-making module for compliance is implemented through a four-layer architecture. Layer A is the compliance state transition strategy layer, which classifies highway traffic laws based on our categorization [26] and breaks them down into basic violation types. For each basic violation type, a transition path from violation state to compliance state is defined, consisting of compliance references and constraints, this layer is predefined. In Layer B, the current basic violation type of the vehicle is determined based on the results of online violation monitoring. Then, using the state transition strategy of Layer A, the transition paths specific to the current basic violation type are determined. Compliance references and constraints are generated based on these transition paths to ensure compliance. It is important to note that the vehicle may simultaneously exhibit multiple violations, so this layer provides multiple compliance references and constraints. Layer C assigns traffic law priorities, given that the final decision algorithm can only apply a set of reference constraints. By prioritizing the Layer B results, the optimal compliance references and constraints can be chosen. Finally, Layer D selects the initial reference trajectory or the compliance reference constraints as the final reference to be passed to the MPC for optimization and tracking based on the results of online violation monitoring.

Through this framework, the vehicle can avoid violating traffic laws by temporarily abandoning its initial trajectory and implementing compliant behavior. This ensures safe and legal highway driving.

## III. COMPLIANCE DECISION-MAKING METHOD

### A. Decomposition of Traffic laws and triggering condition definition

#### 1) Decomposition of Traffic laws:

During vehicle operation, legal actions are nested with each other. For example, the overtaking action includes lane-changing action, and lane-changing includes speed limit and following distance requirements. Therefore, a vehicle may be simultaneously involved in multiple violations at any given time. Considering compliance actions for each individual traffic law would generate a substantial workload and limit scalability. Hence, this paper proposes a method for decomposing traffic laws into fundamental types of violations.

According to research published in [26], despite the fact that traffic laws in different countries and regions are influenced by their local cultural history and social background, these laws share common principles. There is a high degree of similarity between the constraints imposed on the driving behavior of vehicles. For instance, most countries establish speed limits on roads and require drivers to maintain a safe following distance. Furthermore, the expression and meaning of traffic signs and road markings are generally consistent, indicating that the internal logic of traffic laws in various countries is coherent. Through analysis, it has been determined that the traffic laws in most countries and regions primarily restrict driving behavior in the following four aspects: **speed restriction**, **distance restriction**, **behavioral restriction**, and **road right restriction**.

In the following sections of this paper, the Chinese highway traffic laws serve as an illustration of the method proposed in this paper. Since this paper focuses solely on compliant driving on main highway roads, we use the four traffic laws of the *speed limit*, *following distance*, *lane-changing*, and *overtaking* on Chinese highways as examples to introduce the method proposed in this paper. Traffic laws can be found on our website: https://github.com/SOTIFAVLab/DOTL/blob/main/article.pdf.

*Speed limit* belong to the first category of traffic law restrictions, mainly divided into two situations: below the lower speed limit and above the upper speed limit. The speed limit may be either the speed limit established by the road or the speed limit indicated by road signs. As for *following distance*, which belongs to the second category of traffic law restrictions and only exists in one situation, inadequate following distance. Regarding *lane-changing*, they consist of two stages: lane keeping and lane-changing stage. The lane-keeping stage includes *speed limit* and *following distance*, which are the same as the previous analysis. The lane-changing stage belongs to the fourth stage and lane-changing stage. The lane-keeping stage includes *speed limit* and *following distance*, which are the same as the previous analysis. The lane-changing stage belongs to the fourth category of traffic law restrictions, which is road right restriction. In this section, it is simply divided into two situations: left lane unsafe and right lane unsafe. There are three stages of *overtaking*: the left lane change stage, the overtaking stage, and the right lane change stage. The classification of left and right lane change stages is the same as the *lane-changing* laws. Regarding the overtaking phase, it is divided into two violation types based on speed and behavior restrictions. This paper classifies them as follows: exceeding lane line timeout and inadequate overtaking speed. In summary, the basic types of violations on highways are shown in TABLE I.

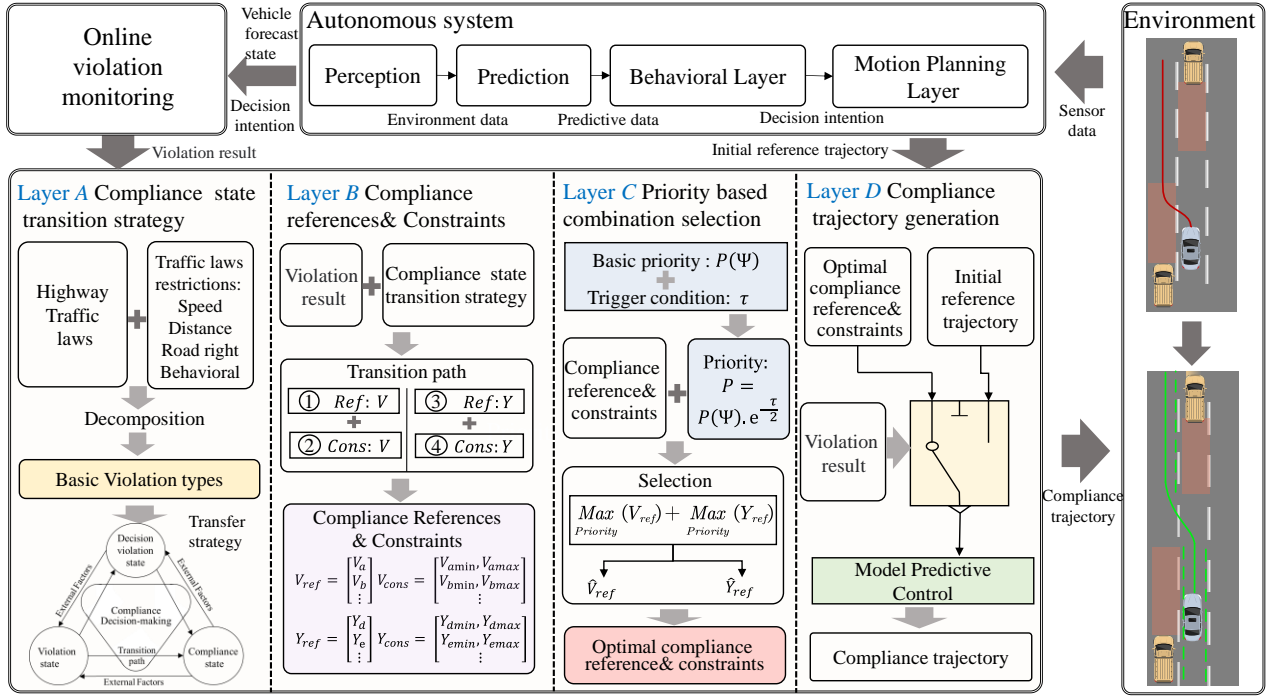

Fig. 2. Trigger-based layered compliance decision-making framework.

TABLE I. Highway basic violation types.

| Basic violation types | Logical expression |
|---|---|
| **a.** Speed below the lower speed limit | $v_{ego} < v_{speed\_min}$ |
| **b.** Speed above the upper speed limit | $v_{ego} > v_{speed\_max}$ |
| **c.** Insufficient front distance | $(vx(Ego) > 100km/h \wedge distance(Ego,Tgt_f) < 100m)$ $\vee (vx(Ego) \leqslant 100km/h \wedge distance(Ego,Tgt_f) < 50m)$ |
| **d.** Left Lane unsafe | $\exists Tgt_{r\text{-}fl} \wedge \binom{TTCX(Ego,Tgt_{r\text{-}fl}) \leqslant TTCX}{\vee distance(Ego,Tgt_{r\text{-}fl}) \leqslant d_{clmin}}$ |
| **e.** Right Lane unsafe | $\exists Tgt_{r\text{-}fr} \wedge \binom{TTCX(Ego,Tgt_{r\text{-}fr}) \leqslant TTCX}{\vee distance(Ego,Tgt_{r\text{-}fr}) \leqslant d_{clmin}}$ |
| **f.** Vehicle override lane line timeout | $(t - t_{in}) > t_{max\_cl}$ |
| **g.** Unsuitable overtaking speed | $v_{ego} < v_{Tgt} + \Delta v_{ot}$ |

Where $v_{ego}$ represent the speed of the ego vehicle, $v_{speed\_min}$ and $v_{speed\_max}$ represent the lower limit and upper limit of the speed allowed by the current lane.

$distance(Ego,Tgt_f)$ represent the distance between the ego vehicle and the vehicle in front of the same lane, $TTCX$ is the time to longitudinal collision, $d_{clmin}$ is the minimum actual distance of a vehicle from relevant vehicles, $t_{in}$ denotes the time when start to drive on the lane line, $t_{max\_cl}$ represents the maximum allowable time for driving on the lane line, $v_{Tgt}$ represent the speed of the target vehicle being overtaken, and $\Delta v_{ot}$ indicates the compliant speed difference during overtaking.

Based on the statistical analysis results from [26], the threshold settings mentioned above in this article are as shown in TABLE II.

TABLE II. Compliance determination threshold.

| Threshold | Value | Unit |
|---|---|---|
| TTCX | 2.3 | s |
| $d_{clmin}$ | 14 | m |
| $t_{max\_cl}$ | 6 | s |
| $\Delta v_{ot}$ | 15 | m/s |

**2) Triggering condition definition:**

Although the applicability conditions for each law are specified in traffic laws, they are typically expressed in ambiguous natural language, making them unsuitable for direct use by AVs. Therefore, it is necessary to define the applicability conditions of traffic laws as trigger conditions through digitization. This paper prioritizes using higher-level decision intentions as trigger signals for the trigger conditions. If it is not possible to obtain higher-level decision intentions, it is necessary to define trigger conditions based on certain vehicle state variables; this paper uses four major highway traffic laws as examples. On highways, vehicles must always adhere to the speed limit and following distance. Therefore, the *speed limit* and *following distance* are continuously monitored and lack trigger conditions. For *lane-changing*, if the vehicle's decision intent can be obtained, the decision intent generated by the behavioral layer serves as the trigger signal. If the decision intent cannot be determined, refer to [26] for the design of the trigger condition. The trigger conditions for left lane-changing are that the ego vehicle decides to change lanes to the left and overlaps with the left lane line. The corresponding mathematical expression is expressed as Eq. (1).

For *overtaking*, like *lane-changing*, if the overtaking intent can be obtained, that intent signal serves as the trigger condition. If the overtaking intent cannot be obtained, the overtaking trigger condition is met when there is a vehicle ahead of the ego vehicle at a short distance (TTC less than 20s) and the leading vehicle is traveling at a slower speed than the ego vehicle. Additionally, the ego vehicle possesses a lateral speed (speed greater than 0.25m/s). The corresponding mathematical expression is as Eq. (2).

$$T_{changeleftlane} \Leftrightarrow overlap\left(Area(Ego),\ y_{line}(lane(ego))\right) \\ \wedge v_y(Ego) > 0 \tag{1}$$

$$T_{overtaking} \Leftrightarrow \\ \begin{pmatrix} \exists Tgt \in front(Tgt, Ego) \wedge v_x(Tgt) < v_x(Ego) \wedge \\ TTCX(Ego, Tgt) < 20s \wedge |v_y(Ego)| > 0.25m/s \end{pmatrix} \tag{2}$$

### B. Compliance constraints and reference generation

In this section, we define the compliance state and violation state for each of the previously identified basic violation types, as well as their respective transition paths. As compliance is achieved through the application of compliance references and constraints, this section defines the compliance references and constraints for each of the basic violation types.

Considering the lateral and longitudinal control requirements of the vehicle, decoupling the lateral and longitudinal control of the ego vehicle is desired. In this paper, longitudinal control is achieved through the reference speed, while lateral control is achieved through the lateral reference position. Therefore, the compliance reference and constraint primarily pertain to the speed $V$ and lateral position $Y$ of the vehicle. This paper defines four fundamental compliance state transition pathways. The state transition paths for basic vehicle violation types are composed of these four basic paths, as shown in TABLE III.

TABLE III. Basic transition path.

| Sign | Basic transition path | Variable |
|---|---|---|
| ① | Compliance reference speed | $V_{ref}$ |
| ② | Compliance speed constraint | $V_{cons}$ |
| ③ | Compliance lateral position reference | $Y_{ref}$ |
| ④ | Compliance lateral position constraint | $Y_{cons}$ |

The next task is to analyze each of the basic violation types and define their compliance references and constraints：

(1) *a*: In **Fig.** 3, the basic violation type *a* is shown to have three vehicle states: violation, when the vehicle speed is below the lower limit required by the traffic laws, decision violation, when the vehicle speed has become legal, but the initial reference speed given by the motion planner is still illegal, and compliance, when both the speed and reference speed are legal. When the vehicle is in a violation state, a suitable compliance reference speed is required to prompt the vehicle to comply immediately; the transition path would be ① . The initial reference speed is truncated and the new compliance reference speed $V_{ref}$, is directly transmitted to the downstream MPC layer. To ensure the vehicle's speed compliance with the legal while maintaining the driving intention as much as possible, $V_{ref}$ is set as the lower limit of the compliance speed.

$$V_{ref}(ego) = V_{speed\_min} \tag{3}$$

Even though the vehicle speed is compliant when the ego vehicle is in the decision violation state, the initial reference speed may cause the vehicle to repeatedly switch between compliance and violation states. To avoid this, the compliance reference speed $V_{ref}$ must replace the initial reference speed and be set to the minimum compliance speed. Additionally, to prevent the vehicle speed from violating the law again, the corresponding compliance speed constraint $V_{cons}$ should be imposed, which is set within the current lane compliance speed range.

$$V_{ref}(ego) = V_{speed\_min} \tag{4}$$

$$V_{cons} \in [V_{speed\_min}, V_{speed\_max}] \tag{5}$$

When the ego vehicle is in a compliance state, both the vehicle speed and the initial reference speed are compliant, and the latter is transmitted directly to the control layer for execution. However, to prevent speed violations, the compliance speed constraint remains.

$$V_{ref}(ego) = Initial\ reference\ speed \tag{6}$$

$$V_{cons} \in [V_{speed\_min}, V_{speed\_max}] \tag{7}$$

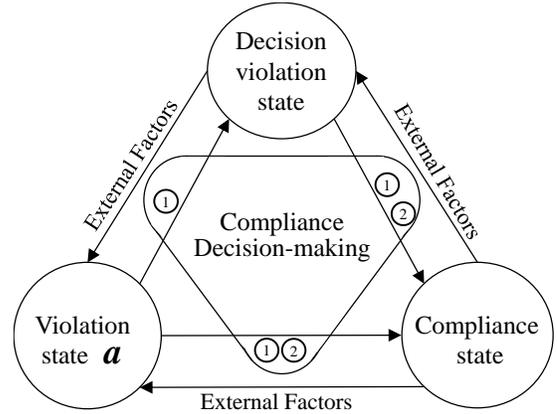

**Fig.** 3. State transfer diagram of *a*.

(2) *b*: The basic violation type *b* is the same as *a*, which also has three states: violation state where the vehicle speed exceeds the upper limit of compliance speed, decision violation state, and compliance state. In the violation state, $V_{ref}$ is set as the upper limit of compliance speed.

$$V_{ref}(ego) = V_{speed\_max} \tag{8}$$

When the ego vehicle is in decision violation state, $V_{ref}$ is also set as the maximal compliance speed, which is the same as Eq. (8). Similarly, speed constraints must be applied to prevent further violations, which is the same as Eq. (7):

When the ego vehicle is in a compliance state, the reference speed is the initial reference speed. The speed constraint is the same as Eq. (7).

(3) $c$: In **Fig.** 4, basic violation type $c$ has two states for the vehicle: violation state, where the distance between the ego vehicle and the front vehicle is non-compliant, and compliance state, where the distance meets the *following distance*. In the violation state, a compliance reference vehicle speed must be provided to the autonomous vehicle in order for it to decelerate as much as possible and maintain a compliant distance from the front vehicle. Additionally, it is against the law to overtake or change lanes when the distance to the front vehicle is insufficient according to traffic laws. To prevent the vehicle from changing lanes, lateral position reference and lane constraints must also be applied. Hence, for the basic violation type $c$, the transition path would be ①,③ *and* ④ . First, for the compliance reference speed, this study assumes a constant rate of deceleration for the ego vehicle, while the speed of the front vehicle remains unchanged during this process. The reference speed is defined as follows:

$$V_{ref}(ego) = -\frac{t_1(V_{0ego} - V_{0Tgt})}{t_2} -$$
$$\frac{2(D - Tgt_{x0} + ego_{x0})}{t_2} + v_{0Tgt} \qquad (9)$$

Where $t_1$ and $t_2$ refer to the time taken to decelerate from the ego vehicle to the same speed as the front vehicle and the time taken to reach the compliance distance between the ego vehicle and the front vehicle, respectively. $V_{0ego}$ and $V_{0Tgt}$ refer to the speed of the ego vehicle and the speed of the front vehicle when the compliance decision is triggered. $D$ represents the compliance distance (when the speed of the vehicle on the highway is greater than 100km/h, D is 100m, and when the speed of the vehicle is lower than 100km/h, D is 50m). $ego_{x0}$ and $Tgt_{x0}$ represent the initial position of the ego vehicle and the front vehicle when the compliance decision is triggered.

Then, for the compliance reference lateral position and lane constraints, the vehicle should travel along the centerline of the current lane when the distance to the front vehicle is insufficient. Therefore, the lateral reference position for compliance is defined as the centerline of the lane. The lane constraints take the position constraint of the vehicle's center of mass into account.

$$Y_{ref} = \binom{y_{lineright}(Initiallane) +}{y_{lineleft}(Initiallane)} / 2$$

$$Y_{cons} \in \binom{y_{lineright}(lane(ego)) + \frac{w_{ego}}{2},}{y_{lineleft}(lane(ego)) - \frac{w_{ego}}{2}} \qquad (10)$$

Where $Y_{ref}$ represents the lateral coordinate of the reference trajectory, *Initiallane* refers to the initial lane where the vehicle is located when the compliance decision-making is triggered, and *lane(ego)* represents the current lane where the vehicle is driving,

$y_{lineright}$ and $y_{lineleft}$ represent the right and left lane lines of the lane, respectively, and $w_{ego}$ refers to the vehicle width.

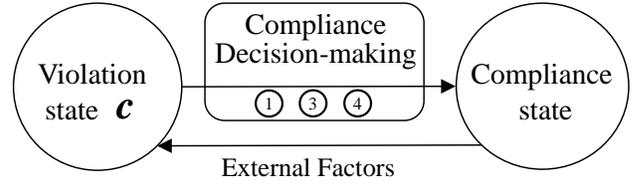

**Fig.** 4. State transfer diagram of $c$.

In the compliance state, the initial reference trajectory is used as the reference trajectory sent to the control layer.

$$Y_{ref} = Initial\ reference\ trajectory \qquad (11)$$

(4) $d$: The vehicle in basic violation type $d$ has two states: violation state and compliance state. The violation state occurs when a vehicle intends to switch lanes to the left, but the left lane is unsafe. In such cases, the vehicle should abandon the left lane change and continue driving in the current lane. Therefore, the transition path from violation state to compliance state is ③,④ , which means that compliance reference lateral position and lane constraints should be applied:

$$Y_{ref} = \binom{y_{lineright}(Initiallane) +}{y_{lineleft}(Initiallane)} / 2 \qquad (12)$$

$$Y_{const} \in \binom{y_{lineright}(lane(ego)) + \frac{w_{ego}}{2},}{y_{lineleft}(lane(ego)) - \frac{w_{ego}}{2}} \qquad (13)$$

When the vehicle is in compliance state, the reference trajectory is the initial reference trajectory, which is same as Eq. (11).

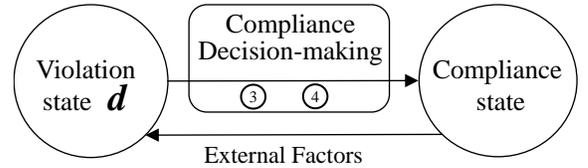

**Fig.** 5. State transfer diagram of $d$.

(5) $e$: Like the state transition diagram for basic violation type $d$, type $e$ also has two states: violation state and compliance state. The violation state occurs when the vehicle intends to change lanes to the right, but the right lane is unsafe. In such cases, the vehicle should temporarily abandon the right lane change and continue driving in the current lane. Therefore, the transition path from violation state to compliance state is ③,④ , indicating that compliance reference lateral position and lane constraints should be applied. Note that in type $e$, there are two situations: when the ego vehicle intends to change lanes to the right, and when the ego vehicle intends to return to its original lane after overtaking. In such situations, the compliance reference trajectory cannot be represented by the initial lane, but by the current lane. The same lane constraints as described in Eq. (13) apply.

$$Y_{ref} = \binom{y_{lineright}(lane(ego)) +}{y_{lineleft}(lane(ego))} / 2 \qquad (14)$$

When the vehicle is in compliance state, the reference trajectory is set as the initial reference trajectory, which is the same as Eq.(11).

(6) **$f$**: For basic violation type $f$, the violation state occurs when the vehicle runs over the lane line for a duration exceeding the specified threshold. The compliance threshold is shown in Table II. According to the state transition diagram in **Fig.**6, when this violation occurs, a compliance reference lateral position should be provided to guide the vehicle to return to the center of the lane as soon as possible. Therefore, the transition path is ③ , and the compliance reference lateral position is defined as the centerline of the lane where the vehicle's center of mass is located, which is the same as in Eq. (14).

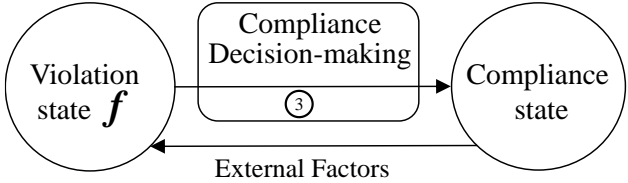

**Fig**. 6 State transfer diagram of $f$

(7) **$g$**: For basic violation type $g$, as shown in **Fig.** 7, a violation occurs when the vehicle attempts to overtake and then returns to its original lane without achieving the required overtaking speed difference. The overtaking speed difference threshold is specified in Table II. In this case, the first step is to provide a compliance reference speed in order to accelerate the vehicle and reach the compliant overtaking speed difference. Additionally, if the overtaking speed difference does not comply with traffic laws, the vehicle must remain in the overtaking lane and is not allowed to return to the original lane. Therefore, compliance reference lateral position and lane restrictions must also be provided. The transition path is ①,③ $and$ ④ . The definitions of compliance references and constraints are as follows.

$$V_{ref} = V_{tgt} + \Delta v_{ot} \qquad (15)$$

$$Y_{ref} = \begin{pmatrix} y_{lineright}(Overtakelane) + \\ y_{lineleft}(Overtakelane) \end{pmatrix}/2 \qquad (16)$$

$$Y_{const} \in \begin{pmatrix} y_{lineright}(Overtakelane) + \dfrac{w_{ego}}{2}, \\ y_{lineleft}(Overtakelane) - \dfrac{w_{ego}}{2} \end{pmatrix} \qquad (17)$$

When the vehicle complies, the reference speed and trajectory are the initial reference speed and the initial reference trajectory.

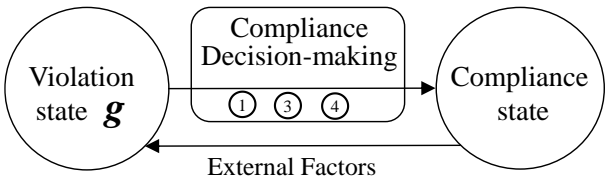

**Fig**. 7. State transfer diagram of **$g$**.

### C. Priority of compliance references and constraints

In the previous section, compliance references and constraints were established for each basic violation type. However, when multiple violations occur simultaneously, conflicts may arise between compliance references and constraints. To avoid this, we assigned appropriate priorities to various types of violations based on an analysis of safety. This paper defines our priorities based on the four categories of traffic law restrictions, taking into account the presence or absence of trigger conditions.

Firstly, in this paper, the four categories of regulatory restrictions are denoted as $\Psi$ , and the basic priority is represented by $P(\Psi)$. However, merely representing the basic priority is insufficient. For instance, even within the same speed restriction category, the generated reference speeds for basic violation types **$a$** and **$g$** may conflict. Therefore, this paper examines the triggering conditions of traffic laws. Traffic laws without trigger conditions are traffic laws that vehicles must always follow and serve as the foundation for safety. On the other hand, traffic laws with trigger conditions, such as lane change and overtaking, can ensure safety by nullifying the decision intention. Therefore, traffic laws with trigger conditions should have a lower priority than those without trigger conditions. In summary, the priority definitions for the traffic laws in this section are as follows:

$$P = P(\Psi).e^{-\frac{\tau}{2}} \qquad (18)$$

Where $\tau$ represents the trigger condition, $\tau = (0, 1)$ , $\tau = 1$ when there is a triggering condition.

For the four categories of traffic laws: **speed restriction**, **distance restriction, road right restriction**, and **behavior restriction**, this paper considers distance restriction to have the highest basic priority. This is because it directly represents the distance between vehicles and is the traffic law that reflects safety the most directly. Therefore, the basic priority of the distance limit is defined as $P(D) = 4$ . Then, in this paper, road right refers to the exclusivity of a certain area for a vehicle at a given moment. A vehicle with a higher road right has the right-of-way over a section of road for a predetermined amount of time. If a vehicle violates the right-of-way and enters this area, the likelihood of a collision is high. Therefore, road right is the second-highest priority and is defined as $P(R) = 3$ . Regarding the remaining speed restriction and behavior restriction, considering safety and comfort, high-speed vehicles are more prone to collisions, and frequent speed changes can also negatively impact comfort. Therefore, speed restriction should have a higher basic priority than behavior restriction. Therefore, the priority definitions for these two are as follows: $P(S) = 2$ , $P(B) = 1$ .

Finally, Eq. (18) can be used to calculate the priority of compliance references and constraints generated for basic violation types. It is essential to note that the priority only comes into play when multiple basic violation types occur simultaneously and the provided compliance references and constraints conflict.

### D. Compliance trajectory generation

MPC is able to receive references and constraints in real-time and perform trajectory optimization[27]. Therefore, in this paper, MPC is adopted as the trajectory optimization and tracking algorithm, which can receive compliance references and constraints provided by the upstream compliance decision-making layer and optimize and track the trajectory accordingly.

Referring to the works mentioned in [28] and [29], the expression for the state space of the vehicle in the global coordinate system in this paper is shown in Eq. (19).

$$\dot{x} = Ax + Bu \qquad (19)$$
$$y = Cx$$

Where:

$$x = [v_x, v_y, \dot{\varphi}, \varphi, X, Y]^T, u = [F_{xT}, \delta_f]^T, y = [v_x, \varphi, Y]^T$$

The cost function consists of the tracking of the reference path and control input. The cost function is as follows.

$$\min_{\Delta u} J = \sum_{k=1}^{N_p} \| y^{t+k,t} - y_{des}^{t+k,t} \|_Q^2 + \sum_{k=1}^{N_c} \| \Delta u_c^{i+k-1,t} \|_R^2 \qquad (20)$$

$$s.t (k = 1, ..., N_p)$$

$$x^{t+k,t} = Ax^{t+k-1,t} + Bu^{t+k-1,t} \qquad (20a)$$

$$y^{t+k,t} = Cx^{t+k,t} \qquad (20b)$$

$$y_{s\_\min}^{t+k,t} \leqslant y_s^{t+k,t} \leqslant y_{s\_\max}^{t+k,t} \qquad (20c)$$

$$u_{c\_\min} < u_c^{t+k-1,t} < u_{c\_\max} \qquad (20d)$$

$$\Delta u_{c\_\min} < \Delta u_c^{t+k,t} < \Delta u_{c\_\max} \qquad (20e)$$

where $N_p$ represents the prediction horizon, $N_c$ is the control horizon, $t + k, t$ represents the predicted value at $k$ steps ahead of $t$, $Q$ is the weighting matrices of the reference, which is $Q = \begin{bmatrix} w_v & 0 & 0 \\ 0 & w_\varphi & 0 \\ 0 & 0 & w_Y \end{bmatrix}$, The $w_v$, $w_\varphi$, $w_Y$ correspond to the weighting parameters for the reference speed, reference heading angle, and reference lateral position, respectively. and $R$ is the weighting matrices of the control input, which is $R = \begin{bmatrix} w_F & 0 \\ 0 & w_\delta \end{bmatrix}$.

The reference trajectory information required by MPC consists of the desired lateral position, longitudinal vehicle speed, and heading angle. In this paper, they are represented by the compliance reference lateral position $Y_{ref}$, compliance reference vehicle speed $V_{ref}$, and a constant reference heading angle of 0, respectively.

$$y_{des} = [v_{xdes}, \varphi_{ref}, Y_{des}]^T = [V_{ref}, 0, Y_{ref}]^T \qquad (21)$$

This paper utilizes MPC's ability to handle multiple constraints, including conventional actuator capacity constraints. Additionally, we introduce compliance speed constraints $V_{cons}$ and compliance lateral position constraints $Y_{cons}$ from Table III as legal constraints. The state constraint equation is as follows:

$$y_s = C_s x \qquad (22)$$

Where the constraints variables $y_s$ are linearized as a function of the inputs as well as states, where $C_s$ denotes the output matrices, $C_s = \begin{bmatrix} 1 & 0 & 0 & 0 & 0 & 0 \\ 0 & 0 & 0 & 0 & 0 & 1 \end{bmatrix}$, here $y_s$ is the hard constraint variables, $y_s = [V, Y]$.

### E. Logical process of legal compliance decision-making

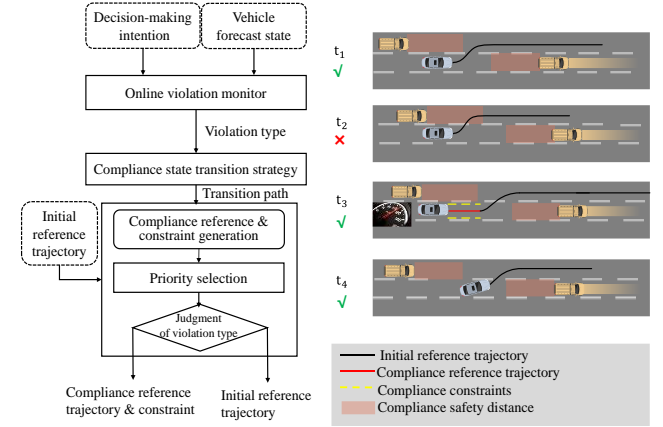

Fig. 8. Logic process of the compliance decision-making.

This section uses the left lane change process as an example to describe the proposed decision-making process for compliance, as shown in **Fig**. 8. The left side illustrates the logical progression of our compliance decision-making, while the right side depicts schematic diagrams at various time points. At $t_1$, the ego vehicle generates a left lane change intention and prepares to change lanes to the left. This decision intention triggers our compliance decision-making module to monitor the violation state of the left lane change process based on the predicted vehicle state. At this moment, the vehicle is compliant, and it follows the initial reference trajectory for driving.

At $t_2$, the vehicle in the left lane suddenly accelerates, and the distance between the ego vehicle and the left rear vehicle is insufficient, indicating that the left lane is unsafe. If the ego vehicle continues to change lanes at this time, it will violate traffic laws, and its speed will be below the minimum speed limit. The online violation monitor identifies $a$ and $d$ as the respective basic violation types. The violation types are then transmitted along with the initial reference trajectory to the state transition strategy module, which uses a predefined state transition diagram to determine the transition paths corresponding to violation types $\boldsymbol{a}$ and $\boldsymbol{d}$. Specifically, for violation type $\boldsymbol{a}$, the compliance reference vehicle speed is provided, while for violation type $\boldsymbol{d}$, the compliance reference lateral position and lateral position constraints are provided. The compliance reference and constraint generation module then generate the corresponding compliance reference and constraint based on the determined transition paths. The generated compliance constraints and references are transmitted to the subsequent module for priority selection. Since the compliance references and constraints for types $\boldsymbol{a}$ and $\boldsymbol{d}$ do not conflict, they are combined into a compliance reference trajectory and transmitted to the downstream MPC layer.

At $t_3$, the ego vehicle drives according to the compliance reference trajectory, which involves temporarily giving up the lane change and accelerating. At $t_4$, the ego vehicle meets the lane change requirements and begins to change lanes according to the initial reference trajectory.

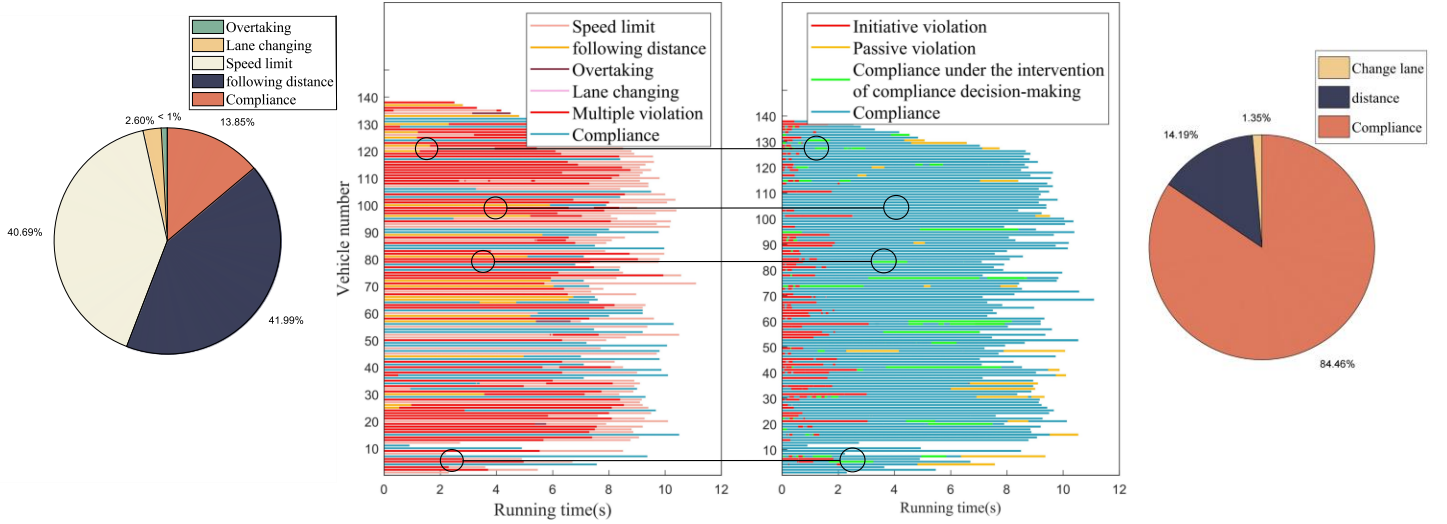

**Fig**. 10. Statistical Analysis of Compliance Decision Pre and Post-AD4CHE dataset.

## IV. EXPERIMENTAL VALIDATION

In this paper, a MATLAB/Simulink model is used to construct the proposed trigger-based layered compliance decision-making framework. MATLAB/Simulink is used to code the online violation monitoring and legal compliance reference and constraint generation methods. To verify the effectiveness of the proposed method, we integrated the compliance decision-making model with the DJI AD4CHE China highway dataset[25] and conducted experimental verification using Simulink-Carsim. The dataset can be obtained from https://auto.dji.com/cn/ad4che-dataset.

### A. The vehicle trajectory dataset

As shown in **Fig**. 9, the AD4CHE dataset provides overhead perspective data captured around Chinese highways. Each vehicle in the dataset is assigned a unique ID based on the order in which it appears. One vehicle is selected as the ego vehicle, while the remaining vehicles in the scene are considered surrounding vehicles. By integrating the dataset into our compliance decision-making model, we first identify violations in the vehicles' behavior through online violation monitoring and then implement compliance decisions to return the vehicles to a compliant state. It is important to note that the AD4CHE dataset is designed for congested road segments, resulting in slower speeds and shorter distances between vehicles, leading to a higher rate of traffic violations. This paper doubles the speeds and relative distances of vehicles in the AD4CHE dataset to increase the dataset's effectiveness. Moreover, the right lane is heavily congested, while the left lane is less congested. Therefore, we only validate our approach to compliance decision-making on the left lane of the AD4CHE dataset. Regarding the triggering conditions and initial reference trajectory, since this paper lacks a behavioral layer, the decision intentions must be approximated using the dataset's data. Among the four traffic laws studied in this paper, the *speed limit and following distance* do not have specific triggering conditions and are always in a monitoring state.

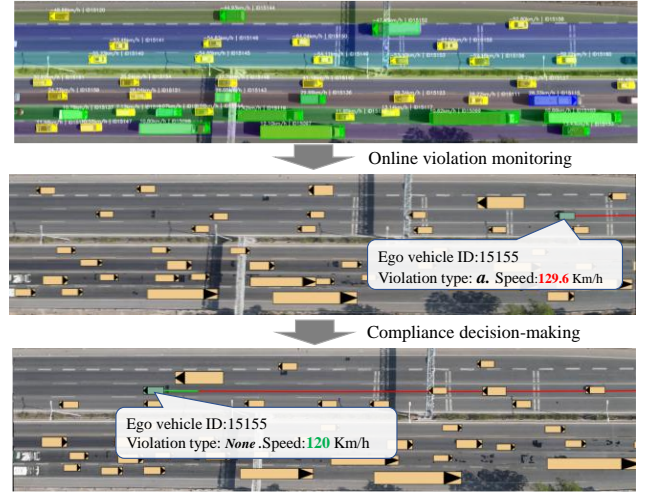

**Fig**. 9. Visualization of the AD4CHE Dataset.

The *lane-changing* and *overtaking* have triggering conditions, namely the lane change intention and overtaking intention. Therefore, a method for approximating decision intentions, as described in [26], is used as shown in Eq. (23)-(25). The initial reference trajectory is obtained from the original trajectories of vehicles in the dataset.

(1) lane-changing intention:

$$\mathrm{v}_y(Ego) > 0.25m/s \Rightarrow$$
$$Decision\_change\_leftlane = True \tag{23}$$

$$\mathrm{v}_y(Ego) < -0.25m/s \Rightarrow$$
$$Decision\_change\_rightlane = True \tag{24}$$

(1) Overtaking intention:

$$\begin{pmatrix} \exists Tgt \in front(Tgt,Ego) \land v_x(Tgt) < v_x(Ego) \land \\ TTCX(Ego,Tgt) < 20s \land |v_y(Ego)| > 0.25m/s \end{pmatrix} \tag{25}$$
$$\Rightarrow Decision\_overtake = True$$

### B. Statistical analysis of results

In this section, we verify the effectiveness of our compliance method by connecting each vehicle in the AD4CHE dataset to the compliance decision-making model. We then traverse the entire dataset to obtain the statistical results. The verification results are shown in **Fig. 10**, and the left figure shows the statistical results obtained by monitoring the real trajectories of all vehicles in the dataset. The horizontal axis represents the time at which the vehicles appear in the dataset, while the vertical axis represents the compliance and violation states of the vehicles throughout their travel. Due to traffic congestion, most vehicles in the dataset are non-compliant most of the time, and the left pie chart clearly shows a compliance rate of only 13.85%. After integrating our compliance decision-making model into the data set, the right statistical chart displays the statistical results of the compliance state of vehicles. It is evident that the violation state is mainly concentrated at the time when the vehicle first appears. This is due to the fact that the vehicles initially appear in the dataset in a violation state. However, with the intervention of our compliance decision-making model, the vehicles quickly transition from a violation state to a compliant state. The green line in the figure, titled Compliance under Compliance Decision-making Intervention, represents a scenario in which the vehicle would violate if it followed the initial reference trajectory, but would remain compliant with the intervention of our compliance decision-making model. Notably, this scenario only occurs during lane-changing and overtaking, as the vehicle's intention to make a decision only arises during these maneuvers. We also observe that there are still passive violations, which are caused by the unreasonable actions of surrounding vehicles. After integrating our compliance decision-making model, the compliance rate of vehicles increased to 84.46%, as shown by the pie chart on the right. Please note that all vehicles in the dataset traveled safely without any collisions. The statistical chart only depicts the compliance or violation state of vehicles.

Comparing the two statistical charts shown in **Fig. 10**, it is evident that our proposed compliance decision-making method can effectively reduce the incidence of driving violations and ensure vehicle compliance.

### 1) Legal compliance verification of speed limit:

As shown in **Fig.11**, for the highway *speed limit*, we chose ego vehicles with IDs 15155 and 15170 from the AD4CHE dataset as validation vehicles.

**Fig. 11** (a) shows the speed profile of vehicle id15155, with the pink curve representing the vehicle's original speed from the dataset. The initial speed of id15155 is consistently non-compliant, as observed. The blue curve represents the vehicle speed after the compliance decision-making model has been incorporated. Initially, both the vehicle speed and initial reference speed are non-compliant within the first second, indicating a violation state. During this time, the vehicle adopts the compliance reference speed $v_{ref}$ provided by our compliance decision-making model. After one second, the vehicle speed becomes compliant. However, the initial reference speed remains non-compliant, indicating a decision violation state. The reference speed is still the compliance reference speed provided by our compliance decision-making module, allowing the vehicle to drive at the maximum speed permitted by the traffic laws.

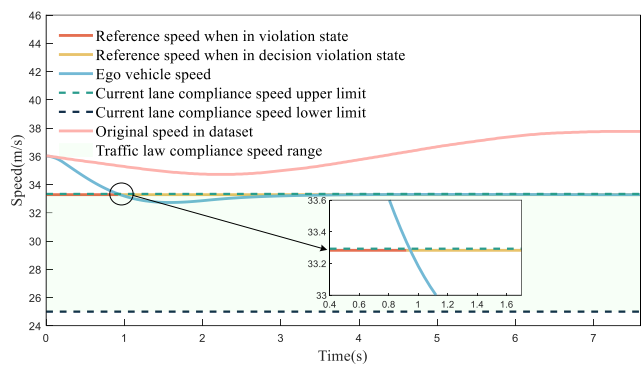

(a) Vehicle speed of id15155

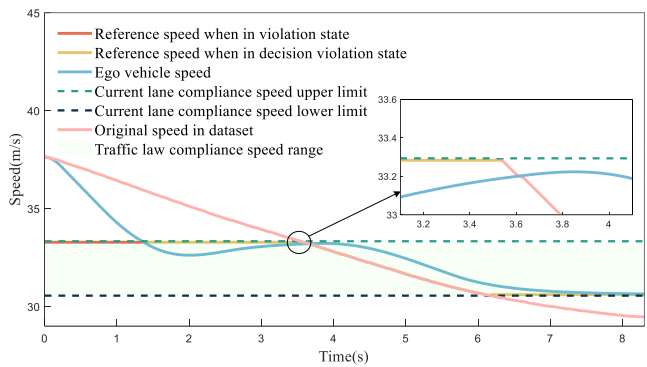

(b) Vehicle speed of id15170 vehicle

**Fig. 11.** Vehicle speed limit verification.

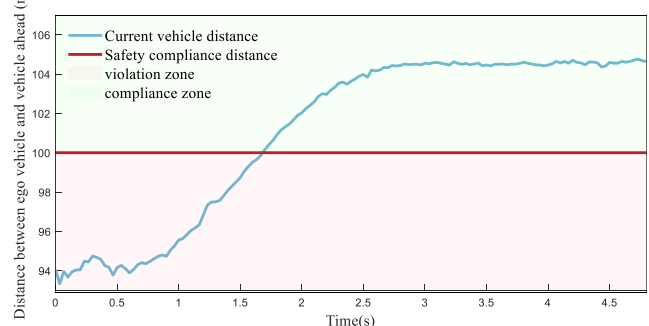

(a) Distance between ego vehicle and the vehicle ahead

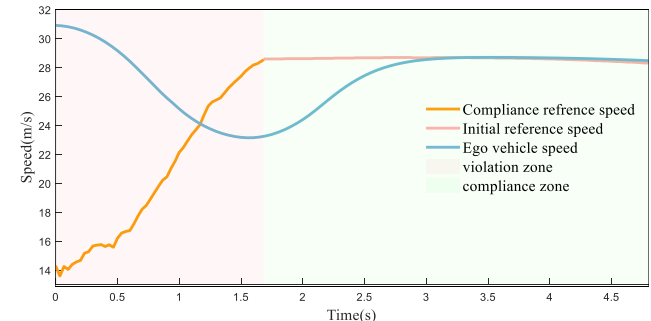

(b) Ego vehicle speed

**Fig. 12.** Following distance verification.

**Fig**. 11(b) shows the speed profile of vehicle id15170. In contrast to id15155, the original speed of id15170 initially exceeds the legal speed limit, then becomes compliant, and eventually drops below the speed limit. After implementing the compliance decision-making model, the vehicle achieves compliance rapidly and maintains high-speed driving. At 3.5s, both the vehicle speed and initial reference speed comply with traffic laws. The reference speed becomes the initial reference speed provided by the motion planner, which is the speed contained in the original dataset. The vehicle begins to decelerate, and at the 6s, the initial reference speed again violates traffic laws. The reference speed becomes our compliance reference speed, and the ego vehicle travels at the lower legal speed limit.

The results indicate that our compliance decision-making method can enable the vehicle to comply as quickly as possible when the speed limit is violated, while preserving the vehicle's initial driving intent as much as possible.

**2) Legal compliance verification of the following distance:**

For *following distance*, when the distance between the ego vehicle and the vehicle ahead is insufficient, we primarily provide the compliance reference speed to maintain the ego vehicle's deceleration at a safe distance from the vehicle ahead, as shown in **Fig.** 12. The selected vehicle is id15129, which starts off less than 100m away from the vehicle in front. The compliance decision module computes a new compliance reference vehicle speed using the current distance, speed, and compliance distance between the leading vehicle and the ego vehicle. The ego vehicle then decelerates according to the compliance reference speed to maintain a safe following distance. To prevent the ego vehicle from oscillating between violation and compliance, this paper sets the compliance distance D to 105m. Once the distance is stable and compliant, the reference speed is reverted to its initial value. Results demonstrate that our compliance decision-making method can rapidly bring a vehicle back into compliance when the following distance is violated.

**3) Legal compliance verification of change lane**

As shown in **Fig.** 13, we selected vehicles with IDs 15150 and 15228 from the AD4CHE dataset as ego vehicles for *lane-changing* validation.

**Fig.**13 (a) shows the trajectory of vehicle id 15150, where the pink curve represents the original trajectory from the dataset. The three images below the trajectory plot are simulation renderings from CarSim, with the blue car representing the ego vehicle and the red cars representing the surrounding vehicles. At time 1, the compliance decision-making model is not triggered as the ego vehicle does not have the intention to change lanes. Therefore, the vehicle adheres to the initial reference trajectory. At time 2, the ego vehicle expresses the intention to switch lanes to the left. However, due to the acceleration of the vehicle in the left rear, the left lane is unsafe, and continuing with the lane change would result in a violation. Therefore, the compliance decision-making model provides compliance references and constraints, causing the ego vehicle to abandon the lane change and continue driving in the current lane. At time 3, the vehicles in the left lane have left, and the left lane is safe. The ego vehicle initiates a lane change.

The id15228 operates similarly to the id15150. At 140 m, it expresses the intention to change lanes, but the lane change would result in a violation. The compliance decision-making model provides compliance references and constraints, causing the ego vehicle to abandon the lane change and continue driving in the current lane. At 230 m, when the left lane becomes safe, the ego vehicle starts to change lanes. Comparing Fig. 13(b)'s magnified view to the original trajectory in the dataset reveals that our compliance decision-making method ensures vehicle compliance while achieving a smoother trajectory.

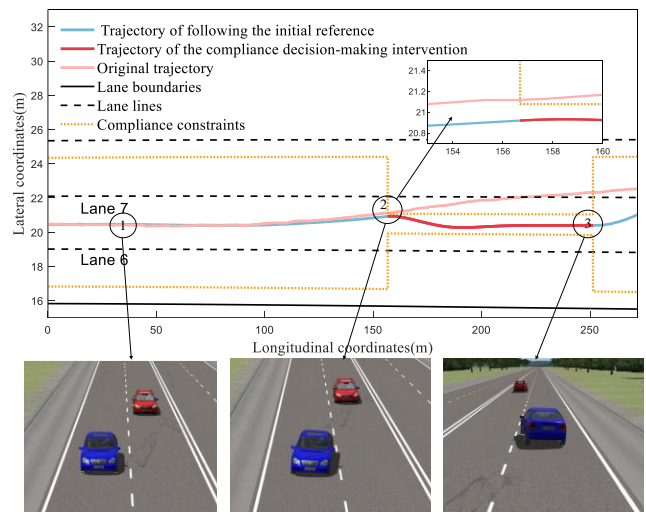

(a) Trajectory of id15150 vehicle

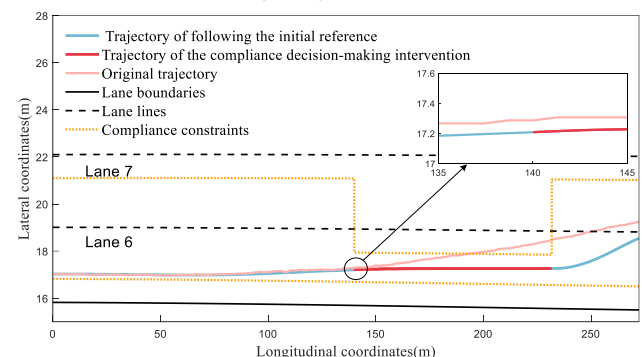

(b) Trajectory of id15228 vehicle
**Fig.** 13. lane-changing verification.

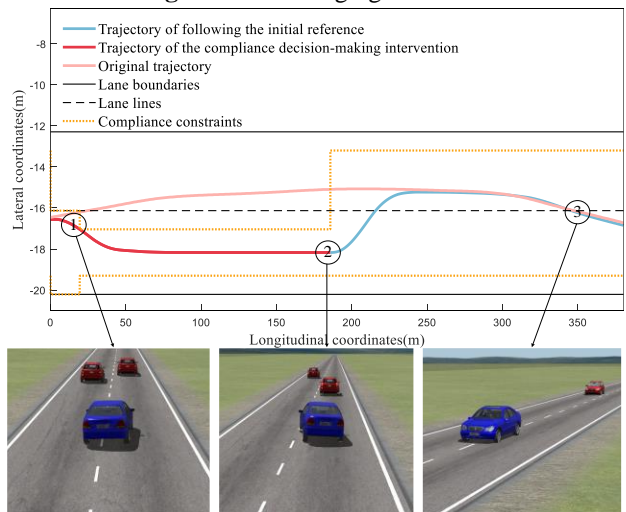

**Fig.** 14. Overtaking verification.

**4) Legal compliance verification of overtaking**

As shown in **Fig.** 14, the pink curve represents the original trajectory from the dataset. The ego vehicle initially intends to overtake, but there is a vehicle in the left lane ahead of it, and the distance between them is unsafe. Therefore, at time 1, the compliance decision-making module provides a compliance reference to cause the ego vehicle to temporarily abandon the overtake and return to the initial lane, while simultaneously imposing compliance constraints that prevent the vehicle from overtaking. At time 2, the vehicle in the left lane has moved away, making the left lane safe, and the ego vehicle starts to overtake. By time 3, the overtaking maneuver is completed, and the vehicle returns to the initial lane.

Based on the above experimental results, it can be concluded that the legal compliance decision-making method proposed in this paper can make the vehicle comply as soon as possible when it violates traffic laws, while maintaining the driver's intent as much as possible. Moreover, when a vehicle is about to violate a traffic law, our method can accurately identify the vehicle's violation intention and generate the corresponding compliance reference trajectory and constraints to maintain the vehicle's compliance.

## V. CONCLUSION

In contrast to existing methods that only consider safety or static non-interactive traffic laws, this paper proposes a decision-making method for determining highway legal compliance that takes into account the compliance of dynamic traffic laws involving interactions with surrounding vehicles while driving. This paper specifically proposes a trigger-based, layered compliance decision-making framework that accurately monitors and decides on vehicle behavior to ensure compliance, using decision intentions from the behavioral layer as trigger signals. By monitoring the predicted state of the vehicle, violation information can be identified before the vehicle violates traffic laws. Based on a four-layer architecture, the compliance decision-making module ensures the scalability and reusability of the method by decomposing traffic laws into basic violation types and formulating compliance state transition strategies. The definition of traffic law priorities ensures that a vehicle prioritizes compliance while maintaining safety in scenarios involving multiple violations. Furthermore, MPC-based trajectory optimization ensures the comfort and smoothness of the compliant trajectory. The validation results on the AD4CHE dataset demonstrate that our method is effective in preventing violations during vehicle operation.

Future research could include the following: (1) Considering the soft and hard constraints of traffic laws to deal with the scenario in which full compliance and security are in conflict; (2) Introducing evaluation indicators to quantify the current violation risk; (3) Considering more complex scenarios to improve the adaptability of the compliance decision-making method; and (4) Considering the real vehicle experiment to validate the timeliness of our method.

## ACKNOWLEDGMENT

We would like to express our gratitude to Professor Yuxin Zhang from Jilin University for providing valuable data support during the experimentation process of this paper.

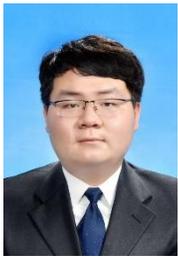

**Xiaohan Ma** received the B.Eng. degree in Vehicle Engineering from the School of Traffic and Transportation of Northeast Forestry University at Harbin, Harbin, China, in 2021. He is currently working toward the MA.Eng. degree in Mechanical Engineering at Beijing Institute of Technology. He is a member of Tsinghua Intelligent Vehicle Design and Safety (IVDAS) Research Institute and supervised by Prof. Jun Li and Hong Wang. His research interests include legal compliance decision-making of autonomous vehicles.

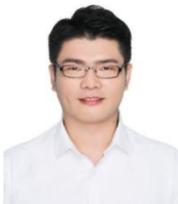

**Wenhao Yu** is currently a Research Associate at Tsinghua University. He received the Ph.D. degree in Transportation Engineering from the School of Automotive and Traffic Engineering, Jiangsu University, China, in 2020. From 2018 to 2019, he was a joint Ph.D. candidate at Mechanical and Mechatronics Engineering, University of Waterloo, Canada. His research focuses on the decision-making, path planning and following control of autonomous vehicles, and safety of the intended functionality of autonomous vehicles, especially the monitor and decision-making algorithms of traffic law compliance.

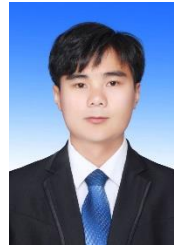

**Chengxiang Zhao** received the B.Eng. degree in Vehicle Engineering from the School of Automotive Engineering of Harbin Institute of Technology at Weihai, Weihai, China, in 2021. He is currently working toward the MA.Eng. degree in Mechanical Engineering at Beijing Institute of Technology. He is a member of Tsinghua Intelligent Vehicle Design and Safety (IVDAS) Research Institute and supervised by Prof. Jun Li and Hong Wang. His research interests include digitization of traffic law.

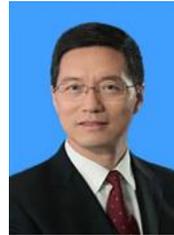

**Changjun Wang** is the Director of Research Institute for Road Safety of the Ministry of Public Security, Beijing, China. He also serves as the Director of National Engineering Laboratory for Road Traffic Optimization and Safety, the President of the China Road Safety Association (CRSA). His research focuses on the technology and engineering of road traffic safety, the development of intelligent transportation system, and the policy analysis of road traffic management.

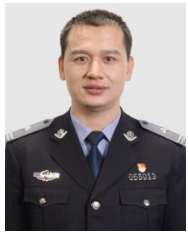

**Wenhui Zhou** is the Director of the Traffic Safety Development Research Department of the Research Institute for Road Safety of the Ministry of Public Security, Beijing, China. Member of the National Technical Committee for Standardization of Road Traffic Management, Member of the National Technical Committee for Standardization of Product Defects and Safety Management, Beijing, China. He has long been engaged in the research of policies, regulations and technologies on vehicle operation safety, road traffic accident investigation, and autonomous vehicle.

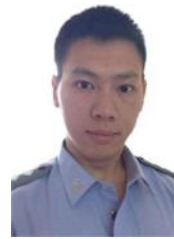

**Guangming Zhao** is Research Associate Professor at the Research Institute for Road Safety of the Ministry of Public Security, Beijing, China. He received the Ph.D. degree in Mechanical Engineering from Wuhan University of Technology, China, in 2013. From 2011 to 2012, he was a joint Ph.D. candidate at Mechanical Engineering, Michigan State University, United States. His research focuses on the policy analysis of vehicle safety management, operation safety technology of autonomous vehicle, especially the decision-making algorithms of road traffic rule compliance and risk response.

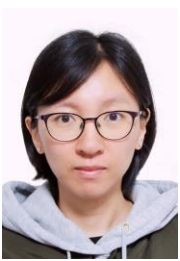

**Mingyue Ma** received the Ph.D. degree with major in transportation engineering from Beihang University, Beijing, China, in 2014. She is an Associate Research Fellow, with Department of motor vehicle safety research, research institute for road safety of MPS, Beijing, China. Her research interests include automatic transmission design, autonomous vehicle safety verification, and vehicle information security.

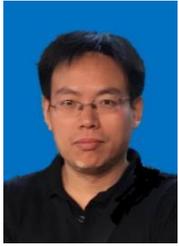

**Weida Wang** received the Ph.D. degree with major in mechanical engineering from Bei hang University, Beijing, China, in 2009. He is currently an Associate Professor in Vehicle Engineering with the School of Mechanical Engineering, Beijing Institute of Technology, Beijing, China. He is the Director of Research Institute of Special Vehicle, Beijing Institute of Technology, Beijing, China. His current research interests include hybrid vehicle, electromechanically transmission control, and energy management strategy.

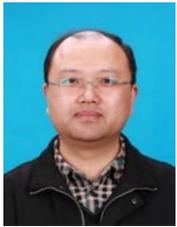

**Lin Yang** received the Ph.D. degree in Mechanical Engineering from Beijing Institute of Technology, Beijing, China, in 2010. Now he is a lecturer of the Beijing Institute of Technology. His current research interests include chassis control and driving assistance systems.

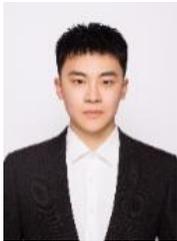

**Rui Mu** received the B.Eng. degree in Transportation Engineering from the School of Transportation and Vehicle Engineering of Shandong University of Technology at Zibo, Shandong, China, in 2021. He is currently working toward the MA.Eng. degree in Transportation Engineering at Jiangsu University. He is a member of Tsinghua Intelligent Vehicle Design and Safety (IVDAS) Research Institute and supervised by Prof. Jun Li and Hong Wang. His research interests include traffic law compliance decision-making of autonomous vehicles.

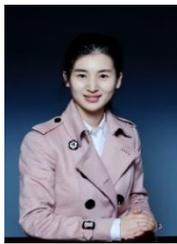

**Hong Wang** is Research Associate Professor at Tsinghua University. She received the Ph.D. degree in Beijing Institute of Technology, Beijing, China, in 2015. From the year 2015 to 2019, she was working as a Research Associate of Mechanical and Mechatronics Engineering with the University of Waterloo. Her research focuses on the safety of the on-board AI algorithm, the safe decision-making for intelligent vehicles, and the test and evaluation of SOTIF. She becomes the IEEE member since the year 2017. She has published over 60 papers on top international journals. Her domestic and foreign academic part-time includes the associate editor for IEEE Transactions on Vehicular Technology and Intelligent Vehicles Symposium, Guest Editor of Special Issues on Intelligent Safety of Automotive Innovation, Young Communication Expert of Engineering, lead Guest Editor of Special Issues on Intelligent Safety of IEEE Intelligent Transportation Systems Magazine.

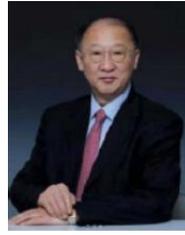

**Jun Li** received the Ph.D. degree in vehicle engineering from Jilin University, Changchun, Jilin, China, in 1989. He is currently an academician of the Chinese Academy of Engineering, a professor at school of Vehicle and Mobility with Tsinghua University, president of the Society of Automotive Engineers of China, director of the expert committee of China Industry Innovation Alliance for the Intelligent and Connected Vehicles. His research interests include internal combustion engine, electric drive systems, electric vehicles, intelligent vehicles, and connected vehicles.